\documentclass[10pt,twocolumn,letterpaper]{article}

\usepackage{cvpr}
\usepackage{times}
\usepackage{epsfig}
\usepackage{graphicx}
\usepackage{amsmath}
\usepackage{amssymb}
\usepackage[symbol]{footmisc}
\usepackage[font=small,labelfont=bf]{caption}
\usepackage{subcaption}

\usepackage{array}
\newcolumntype{?}{!{\vrule width 1pt}}
\cvprfinalcopy 

\def\cvprPaperID{77} 

\begin{document}

\renewcommand{\thefootnote}{\fnsymbol{footnote}}
\newcommand{\superscript}[1]{\ensuremath{^{\textrm{#1}}}}

\title{Feature Forwarding for Efficient Single Image Dehazing}
\author{
    Peter Morales\superscript{*} \space\space\space\space Tzofi Klinghoffer\superscript{*} \space\space\space\space Seung Jae Lee\\
    MIT Lincoln Laboratory\\
    Lexington, MA 02421\\
    {\tt\small \{Peter.Morales,Tzofi.Klinghoffer,SeungJae.Lee\}@ll.mit.edu}
    \vspace{-20pt} 
}

\twocolumn[{
\renewcommand\twocolumn[1][]{#1}\
\maketitle
\setlength\tabcolsep{2pt}
\begin{center}
    \centering
    \begin{tabular}{ccc?cc?ccc}
    \includegraphics[width = 1.5in, height = 0.75in, keepaspectratio]{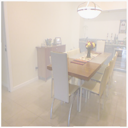} &
    \includegraphics[width = 1.5in, height = 0.75in, keepaspectratio]{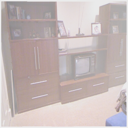} & 
    \includegraphics[width = 1.5in, height = 0.75in, keepaspectratio]{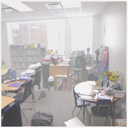} &
    \includegraphics[width = 1.5in, height = 0.75in, keepaspectratio]{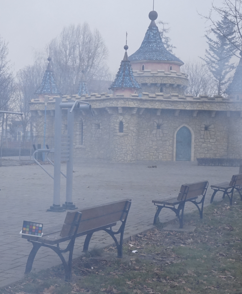} & 
    \includegraphics[width = 1.5in, height = 0.75in, keepaspectratio]{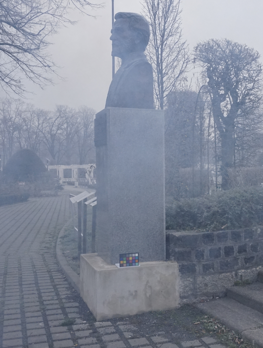} &
    \includegraphics[width = 1.5in, height = 0.75in, keepaspectratio]{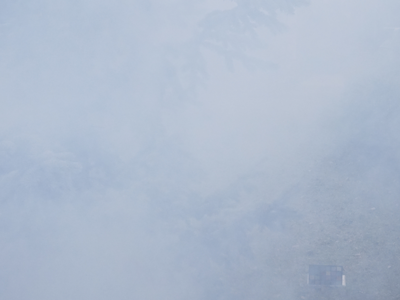} &
    \includegraphics[width = 1.5in, height = 0.75in, keepaspectratio]{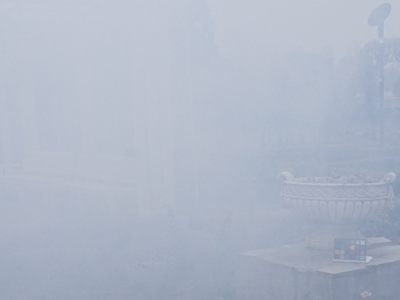} &
    \includegraphics[width = 1.5in, height = 0.75in, keepaspectratio]{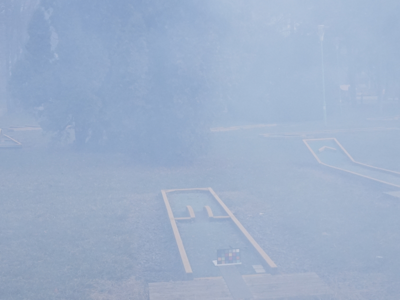}
    \\
    \includegraphics[width = 1.5in, height = 0.75in, keepaspectratio]{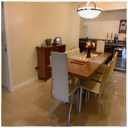} &
    \includegraphics[width = 1.5in, height = 0.75in, keepaspectratio]{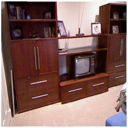} & 
    \includegraphics[width = 1.5in, height = 0.75in, keepaspectratio]{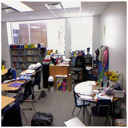} & 
    \includegraphics[width = 1.5in, height = 0.75in, keepaspectratio]{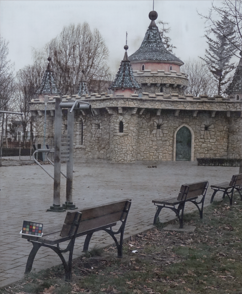} & 
    \includegraphics[width = 1.5in, height = 0.75in, keepaspectratio]{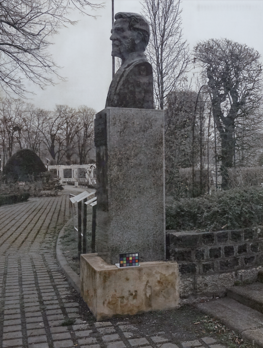} &
    \includegraphics[width = 1.5in, height = 0.75in, keepaspectratio]{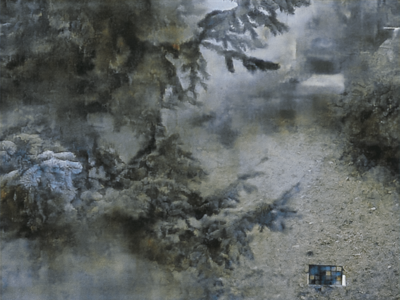} &
    \includegraphics[width = 1.5in, height = 0.75in, keepaspectratio]{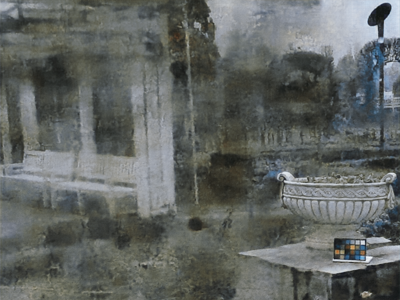} &
    \includegraphics[width = 1.5in, height = 0.75in, keepaspectratio]{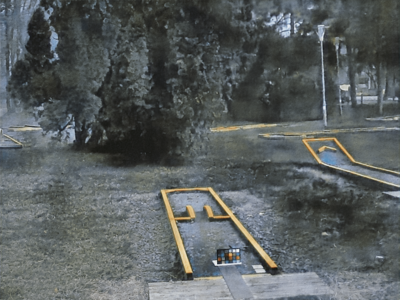} 
    \\
    \multicolumn{3}{c}{NYU Depth} &  \multicolumn{2}{c}{O/I-HAZE} & \multicolumn{3}{c}{Dense-Haze}
    \end{tabular}
    \vspace{-10pt}  
    \captionof{figure}{(top) Hazy images. (bottom) Images after haze is removed using our fully convolutional neural network (CNN) approach.}
    \label{figure:preview}
\end{center}
}]
\maketitle
{\let\thefootnote\relax\footnotetext{\superscript{*} Equal contribution.}}

\setlength{\abovedisplayskip}{2pt}
\setlength{\belowdisplayskip}{2pt}

\vspace{-12pt}
\begin{abstract} 
	Haze degrades content and obscures information of images, which can negatively impact vision-based decision-making in real-time systems. In this paper, we propose an efficient fully convolutional neural network (CNN) image dehazing method designed to run on edge graphical processing units (GPUs). We utilize three variants of our architecture to explore the dependency of dehazed image quality on parameter count and model design. The first two variants presented, a small and big version, make use of a single efficient encoder--decoder convolutional feature extractor. The final variant utilizes a pair of encoder--decoders for atmospheric light and transmission map estimation. Each variant ends with an image refinement pyramid pooling network to form the final dehazed image. For the big variant of the single-encoder network, we demonstrate state-of-the-art performance on the NYU Depth dataset. For the small variant, we maintain competitive performance on the super-resolution O/I-HAZE datasets without the need for image cropping. Finally, we examine some challenges presented by the Dense-Haze dataset when leveraging CNN architectures for dehazing of dense haze imagery and examine the impact of loss function selection on image quality. Benchmarks are included to show the feasibility of introducing this approach into real-time systems.  
\end{abstract}

\section{Introduction}
Many computer vision applications, such as those for image classification and object detection, are trained on datasets comprised of mostly pristine imagery. However, to ensure dependability in real-world environments, computer vision algorithms must be able to perform consistently in various levels of visual degradation. One primary source of image degradation is haze, which introduces challenging, nonlinear noise to a scene. Haze is caused by particulates in the atmosphere, such as dust, fumes, and mist, that absorb and scatter light. Image degradation from haze can adversely affect computer vision algorithms,  making it a principle concern for future systems that incorporate visual information into their decision-making processes.  Previous works \cite{Li_2017_ICCV,DBLP:journals/corr/LinDGHHB16} have established the negative impact of haze on object detection and recognition tasks and have furthermore shown the benefit of introducing image dehazing as a prepossessing step to computer vision tasks. Introducing image enhancement algorithms such as image dehazing may prove to be an important step in creating reliable vision-based systems. 
\subsection{Single Image Dehazing}
The presence of haze in images is often described by the atmospheric scattering model \cite{Koschmieder,middleton1952vision,narasimhan2000chromatic,narasimhan2002vision}, which is classically formulated:
\begin{equation} \label{atom}
I(x) = J(x)t(x) + A(1-t(x))
\end{equation}
where \(I(x)\) is the captured hazy image, \(J(x)\) is the haze-free image, \(A\) is the global atmospheric light, and \(t(x)\) is the transmission map. Consequently, by estimating the global atmospheric light and transmission map for a captured hazy image, the haze-free image can be recovered. This approach has been the basis of several successful approaches \cite{NonLocalImageDehazing,Fattal:2014:DUC:2702692.2651362,he2011single,JIANG201717,JU2017180,6751186,zhu2015fast}. More recently, neural network approaches have also been proposed to estimate these scene properties \cite{dehaze_zhang_2018}.\par
To evaluate the performance of these algorithms, peak signal-to-noise ratio (PSNR) and structural similarity index (SSIM) are commonly used to quantify dehazed image restoration quality. PSNR (measured in decibels) is an absolute error, calculated using the mean square error (MSE) of a pixel relative to its maximum possible value. Alternatively, SSIM attempts to improve upon absolute error metrics by more closely aligning with human perception under the assumption that humans' visual systems are highly attuned to extracting structural information \cite{Wang:2004:IQA:2319031.2320551}. Nevertheless, these metrics do not always agree with human assessment of the similarity of images \cite{ma2015perceptual} and qualitative assessment remains an important component in evaluating performance.
\subsection{Contributions}
In this paper, we present a family of fully convolutional neural network architectures for single image dehazing capable of being deployed on edge GPUs. First, we present two network variants, dubbed Small and Big FastNet, where Small and Big refer to the widths of the networks. Second, we present a neural network based on the atmospheric scattering model that estimates the transmission map and atmospheric light of a scene. We utilize these networks to study change in accuracy as a function of total network parameters, as well as to assess the benefits of estimating a scene's transmission map and atmospheric light. For this paper, we loosely define efficiency based on model performance versus parameter count. All of the proposed networks utilize both an encoder--decoder structure adapted from efficient image segmentation networks \cite{chaurasia2017linknet} and a fully connected pyramid pooling network \cite{DBLP:journals/corr/HeZR014} for output image refinement. Finally, we show benchmarks on reference hardware for varying pixel counts to examine the feasibility of incorporating these algorithms in real-time systems. \par   
The paper makes the following contributions:
\begin{itemize}
  \setlength{\itemsep}{0mm}
  \item A novel neural network architecture that efficiently achieves state-of-the-art performance in single image dehazing on the NYU Depth dataset.
  \item A scaled-down architecture capable of running on super-resolution imagery without the need for cropping, which is a common requirement for previous approaches.
  \item An empirical evaluation of the impact of loss function on restoration quality. 
  \item A discussion of the value of utilizing the atmospheric scattering model when designing neural network image dehazing models.
  \item A discussion on the challenges of using deep learning methods for haze removal, such as the effects from overfitting.
  \item Timing benchmarks for running our architectures on desktop and edge GPUs. 
\end{itemize}
\section{Related Work}
Although there has and continues to be a tremendous amount of success in single image dehazing without the use of neural networks, many recent state-of-the-art techniques utilize deep learning frameworks \cite{DBLP:journals/corr/CaiXJQT16,engin2018cycle,Li_2017_ICCV,dehaze_zhang_2018}. These approaches generally incorporate neural network building blocks originally proposed for image segmentation, style transfer, object detection, and other computer vision tasks. For example, U-Nets \cite{DBLP:journals/corr/RonnebergerFB15}, feature pyramid networks \cite{DBLP:journals/corr/LinDGHHB16}, and residual networks \cite{he2016deep} were all utilized as part of the 2018 NTIRE Image Dehazing Challenge \cite{ancuti2018ntire}. \par
\subsection{Atmospheric Model Learning}
Several successful techniques leverage hand-engineered features to estimate the transmission map for image dehazing \cite{Fattal:2014:DUC:2702692.2651362,he2011single,4587643}. In contrast to these approaches, Cai \etal\cite{DBLP:journals/corr/CaiXJQT16} proposed an end-to-end network that learns features useful for estimating a transmission map. However, this method and similar transmission estimation methods \cite{ren2016single} do not address estimating the atmospheric light within a scene.  Zhang and Patel \cite{dehaze_zhang_2018} addressed this issue by estimating both the atmospheric light and transmission map within a generative adversarial learning framework. In this approach, the unknown variables from the atmospheric scattering model are estimated using independent neural network architectures; U-Net is used to learn atmospheric light and a densely connected network is used to learn a transmission map estimation. Additionally, Li \etal \cite{Li_2017_ICCV} showed that the atmospheric scattering model, described in Equation \ref{atom}, could be reformulated via a linear transform to a single variable and bias. 
\begin{equation} \label{linear}
J(x)=K(x) I(x)-K(x)+b
\end{equation}
\begin{equation}\label{transform}
K(x)=\frac{\frac{1}{t(x)}(I(x)-A)+(A-b)}{I(x)-1}
\end{equation}
This formulation fits naturally within a deep learning framework and hints at the effectiveness of purely convolutional approaches.    
\subsection{Style Transfer and Segmentation Networks}
Generative adversarial networks (GANs) for image style transfer have become increasingly popular in recent years with algorithms such as Pix2Pix \cite{pix2pix2017} and CycleGAN \cite{CycleGAN2017}. Haze removal can also be thought of from a style transfer perspective: transferring images from the hazy domain to the haze-free domain. This approach was attempted by Engin \etal \cite{engin2018cycle}, in which cycle consistency and perceptual losses were combined in a CycleGAN framework.\par
Additionally, approaches from semantic image segmentation, such as feature pyramid networks,  have proven to be effective in image dehazing applications. Image segmentation networks often utilize encoder--decoder pairs to learn embedded representations of inputs that take into account multi-scale features. Chaurasia and Culurciello \cite{chaurasia2017linknet} proposed an efficient semantic segmentation architecture based on a fully convolutional encoder--decoder framework. Their encoder uses a ResNet18 model \cite{he2016deep} for feature encoding and avoids a loss of spatial information by reintroducing residuals from each encoder to the output of its corresponding decoder.\par
\subsection{Super-Resolution Imagery}
One challenge in using neural networks for single image dehazing is processing high-resolution input. Several techniques in the 2018 NTIRE Image Dehazing Challenge handled the relatively high-input resolution of the I-HAZE \cite{I-HAZE_2018} and O-HAZE \cite{O-HAZE_2018} datasets by cropping input imagery into many smaller frames or downsampling the input imagery and resizing the final outputs \cite{ancuti2018ntire}. These approaches are limited by total GPU memory and not GPU processing power; therefore, models with fewer parameters are capable of accepting higher-resolution input imagery. \par 
\section{Proposed Method}
\subsection{Network Architecture}
\begin{figure}[t] 
\begin{center}
   \includegraphics[width=1\linewidth]{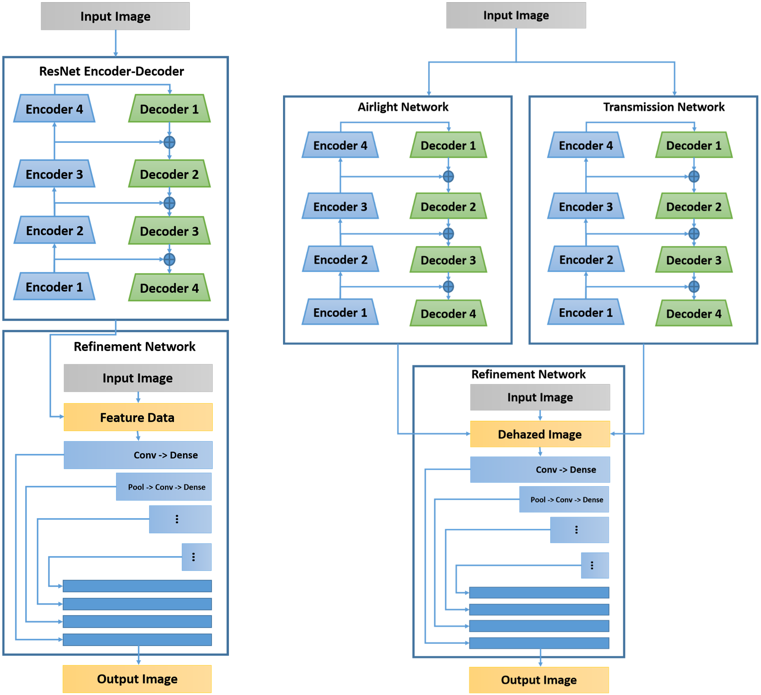}
\end{center}
   \caption{Proposed models: (left) Our FastNet single encoder--decoder architecture, which forwards features directly into a pyramid refinement network. (right) Our DualFastNet architecture, which estimates both atmospheric light and transmission maps to form dehazed images via Equation \ref{atom}.}
\label{fig:long}
\label{fig:onecol}
\label{fig:models}
\label{models}
\end{figure}
Our proposed fully convolutional neural networks (CNNs) build upon past work in efficient image segmentation and deep learning-based image dehazing. For our Small FastNet model, we adapted the LinkNet architecture \cite{chaurasia2017linknet} by removing the final softmax and prediction layers in order to pass features directly into a pyramid pooling network at the full input spatial resolution. LinkNet uses layers from a pretrained ResNet18 model for its encoder modules. For our Big FastNet model, we modified the original architecture's encoder to utilize ResNet50 as its encoder module; we observe that the increased model width (achieved with the deeper ResNet encoder) leads to improved restoration quality at a small speed trade off. Both these models use a single encoder--decoder to learn features of the image, followed by an image refinement pyramid pooling network. The pyramid pooling network helps preserve multi-scale features when forming the final output image by progressively embedding inputs at multiple scales and then resizing all scaled embeddings to the output resolution. \par
In addition to the two single-encoder models, we introduce DualFastNet, which is inspired by past work in atmospheric model networks, notably by Zhang and Patel \cite{dehaze_zhang_2018}. Rather than using a single encoder--decoder, our DualFastNet approach uses two separate encoder--decoder models to learn atmospheric light and transmission map estimations. These estimations are then used as input to calculate a dehazed image using the formulation described in Equation \ref{atom}. This approach was used in our submission to the 2019 NTIRE Image Dehazing Challenge; however, as described in later sections, further studies indicate that Big FastNet yields better performance on larger datasets. Our single encoder--decoder FastNet variant and double encoder--decoder DualFastNet variant are both shown in Figure \ref{models}.\par

\subsection{Implementation and Training Details}
We utilized several loss functions and data augmentation techniques described further in subsequent sections. Our implementation was developed in PyTorch and all results can be generated using our provided code\footnote{https://github.com/pmm09c/ntire-dehazing}. We utilized ADAM \cite{kingma2014adam} as an optimizer for training with an initial learning rate of $1\times 10^{-3}$. During training, validation was done per epoch and models with improved validation loss were saved. Early stopping was used and training ended upon reaching convergence in validation loss to prevent overfitting. Each model was initially trained using MSE as the loss function. However, as described in later sections, some models were fine tuned using a secondary loss function. When validating models based on SSIM and PSNR, we chose to report the model with the highest SSIM, even if the corresponding PSNR was not the highest of all models trained. This means that for all results presented in this paper, models with higher PSNR may be achievable, but with degradation to SSIM.

\section{Experiments}
\subsection{Datasets}
We trained and evaluated our proposed methods on four datasets. First, we leveraged the NYU Depth dataset V2 as prepared by Zhang and Patel \cite{dehaze_zhang_2018}\footnote{https://github.com/hezhangsprinter/DCPDN} and demonstrate an improvement over previous state-of-the-art approaches. This dataset contains 1,000 unique training examples from the NYU Depth dataset V2 \cite{silberman2012indoor} and 4,000 total training samples (each sample has four variations with varying levels of haze). These images are synthesized with the following parameters: \(A\) $\in[0.5,1.0]$ and $\beta\in[1.4,1.6]$, where \(A\) is atmospheric light and $\beta$ is the scattering coefficient. Each training sample consists of a hazy image, an atmospheric light image, a transmission map image, and a dehazed ground truth image. Four hundred test examples are generated in a similar fashion from the NYU Depth dataset V2.\par 
Additionally, we evaluated our approach on the more challenging Dense-Haze dataset \cite{ancuti2019dense} and the high-resolution O- and I-HAZE datasets. These datasets provide real-world imagery that can be used to evaluate the generalizability of our models, the usability of our models on high-resolution data, and the overall performance of our models in various conditions. For the NTIRE 2019 Image Dehazing Challenge, our models were trained exclusively on the Dense-Haze dataset with randomly initialized weights. Models used to evaluate the O/I-HAZE datasets were trained on O/I-HAZE data using weights generated from training the NYU Depth dataset.\par

\subsection{Architecture Comparison}
We studied three variants of our fully convolutional neural network: Small FastNet, Big FastNet, and DualFastNet. As a result of studying these models, we present empirical evidence of the benefits of increasing model width and show the capability of fully convolutional methods to generalize image dehazing mechanisms without the need for an explicit atmospheric model. In later sections, we use the Dense-Haze dataset to show that introducing model priors through the atmospheric scattering model, as done in the DualFastNet architecture, can benefit training when limited training samples are available. \par
Each model was trained and tested on the NYU Depth dataset with MSE loss only and we report the resulting PSNR and SSIM. MSE loss was enforced on the output image of both Small and Big FastNet. For DualFastNet, we examined three ways to train our model. Originally proposed by Zhang and Patel \cite{dehaze_zhang_2018}, we employed a stage-wise learning technique to train atmospheric light, transmission map, and image formation networks separately to quicken convergence; training was completed with the entire model being fine tuned. Although this approach was found to be effective, it burdens the training process. We denote this step-wise learning technique as DualFastNet$_{step}$. We also explored whether our DualFastNet model can be trained wholly from scratch --- both with MSE loss enforced on atmospheric light, the transmission map, the dehazed image, and refined output image (DualFastNet$_{MSE\times 4}$), and with MSE loss enforced only on the refined output image (DualFastNet$_{MSE\times 1}$). Results are summarized in Table \ref{table:losses}. 
\begin{table}[t]
\centering
\begin{tabular}{p{0.30\linewidth}p{0.10\linewidth}p{0.10\linewidth}p{0.27\linewidth}}
\hline
 & PSNR & SSIM & Parameters\\
\hline
\hline
Small FastNet & 24.08 & 0.9034 & \textbf{11,554,167}\\
$\text{DualFastNet}_{\text{MSE}\times 1}$ & 24.69 & 0.9081 & 23,072,725\\
$\text{DualFastNet}_{\text{MSE}\times 4}$ & 22.30 & 0.8650 & 23,072,725\\
$\text{DualFastNet}_{\text{step}}$ & 28.13 & 0.9483 & 23,072,725\\
\textbf{Big FastNet} & \textbf{29.69} & \textbf{0.9563} & 28,782,647\\
\hline
\end{tabular}
\caption{We compare the three variations of the proposed model architecture. Each model is trained until convergence with MSE loss only. DualFastNet is trained with three methods.  Models were trained and evaluated on the NYU Depth dataset.}
\label{table:losses}
\end{table}
Model performance and parameter count appear to be related; models with higher parameter count yield higher performance. In addition, the step-wise learning technique is the most effective for training the atmospheric scattering model-based DualFastNet. The widest architecture, Big FastNet, performs the best of our proposed architectures in both PSNR and SSIM, indicating that using a wider network is a viable alternative to incorporating an atmospheric model prior into the neural network architecture.


\subsection{Loss Functions and Fine Tuning}
\begin{figure*}
\centering
    \setlength\tabcolsep{2pt}
	\begin{tabular}{cccccccc}
    	\includegraphics[width = 1.5in, height = 0.8in, keepaspectratio]{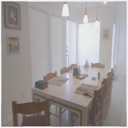} &
    	\includegraphics[width = 1.5in, height = 0.8in, keepaspectratio]{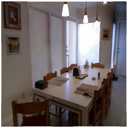} & 
    	\includegraphics[width = 1.5in, height = 0.8in, keepaspectratio]{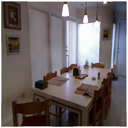} &
    	\includegraphics[width = 1.5in, height = 0.8in, keepaspectratio]{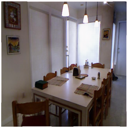} & 
    	\includegraphics[width = 1.5in, height = 0.8in, keepaspectratio]{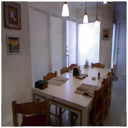} &
        \includegraphics[width = 1.5in, height = 0.8in, keepaspectratio]{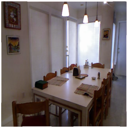} &
        \includegraphics[width = 1.5in, height = 0.8in, keepaspectratio]{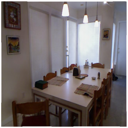} &
    	\includegraphics[width = 1.5in, height = 0.8in, keepaspectratio]{cvpr2019AuthorKit/latex/nyu_depth/gt/6.png}
    	\\
        \includegraphics[width = 1.5in, height = 0.8in, keepaspectratio]{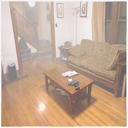} &
    	\includegraphics[width = 1.5in, height = 0.8in, keepaspectratio]{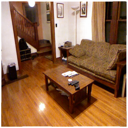} & 
    	\includegraphics[width = 1.5in, height = 0.8in, keepaspectratio]{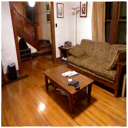} &
    	\includegraphics[width = 1.5in, height = 0.8in, keepaspectratio]{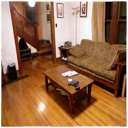} & 
    	\includegraphics[width = 1.5in, height = 0.8in, keepaspectratio]{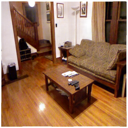} &
        \includegraphics[width = 1.5in, height = 0.8in, keepaspectratio]{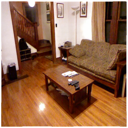} &
        \includegraphics[width = 1.5in, height = 0.8in, keepaspectratio]{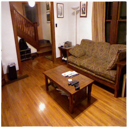} &
    	\includegraphics[width = 1.5in, height = 0.8in, keepaspectratio]{cvpr2019AuthorKit/latex/nyu_depth/gt/67.png}
        \\
        \includegraphics[width = 1.5in, height = 0.8in, keepaspectratio]{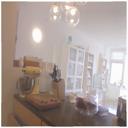} &
    	\includegraphics[width = 1.5in, height = 0.8in, keepaspectratio]{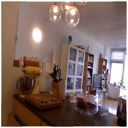} & 
    	\includegraphics[width = 1.5in, height = 0.8in, keepaspectratio]{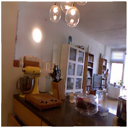} &
    	\includegraphics[width = 1.5in, height = 0.8in, keepaspectratio]{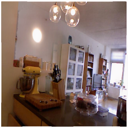} & 
    	\includegraphics[width = 1.5in, height = 0.8in, keepaspectratio]{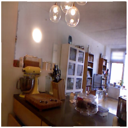} &
        \includegraphics[width = 1.5in, height = 0.8in, keepaspectratio]{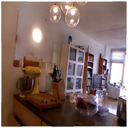} &
        \includegraphics[width = 1.5in, height = 0.8in, keepaspectratio]{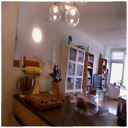} &
    	\includegraphics[width = 1.5in, height = 0.8in, keepaspectratio]{cvpr2019AuthorKit/latex/nyu_depth/gt/209.png}
        \\
        \includegraphics[width = 1.5in, height = 0.8in, keepaspectratio]{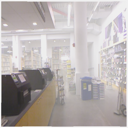} &
    	\includegraphics[width = 1.5in, height = 0.8in, keepaspectratio]{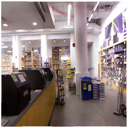} & 
    	\includegraphics[width = 1.5in, height = 0.8in, keepaspectratio]{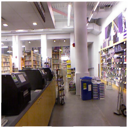} &
    	\includegraphics[width = 1.5in, height = 0.8in, keepaspectratio]{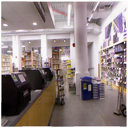} & 
    	\includegraphics[width = 1.5in, height = 0.8in, keepaspectratio]{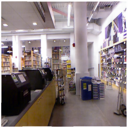} &
        \includegraphics[width = 1.5in, height = 0.8in, keepaspectratio]{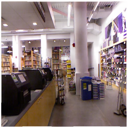} &
        \includegraphics[width = 1.5in, height = 0.8in, keepaspectratio]{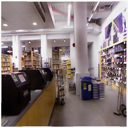} &
    	\includegraphics[width = 1.5in, height = 0.8in, keepaspectratio]{cvpr2019AuthorKit/latex/nyu_depth/gt/291.png}
        \\
        \includegraphics[width = 1.5in, height = 0.8in, keepaspectratio]{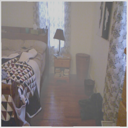} &
    	\includegraphics[width = 1.5in, height = 0.8in, keepaspectratio]{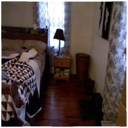} & 
    	\includegraphics[width = 1.5in, height = 0.8in, keepaspectratio]{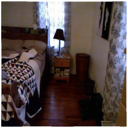} &
    	\includegraphics[width = 1.5in, height = 0.8in, keepaspectratio]{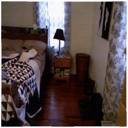} & 
    	\includegraphics[width = 1.5in, height = 0.8in, keepaspectratio]{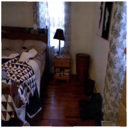} &
        \includegraphics[width = 1.5in, height = 0.8in, keepaspectratio]{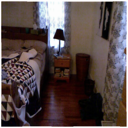} &
        \includegraphics[width = 1.5in, height = 0.8in, keepaspectratio]{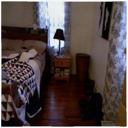} &
    	\includegraphics[width = 1.5in, height = 0.8in, keepaspectratio]{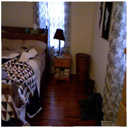}
        \\
    	\includegraphics[width = 1.5in, height = 0.8in, keepaspectratio]{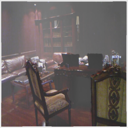} &
    	\includegraphics[width = 1.5in, height = 0.8in, keepaspectratio]{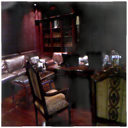} & 
    	\includegraphics[width = 1.5in, height = 0.8in, keepaspectratio]{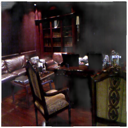} &
    	\includegraphics[width = 1.5in, height = 0.8in, keepaspectratio]{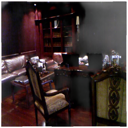} & 
    	\includegraphics[width = 1.5in, height = 0.8in, keepaspectratio]{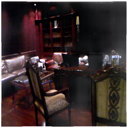} &
        \includegraphics[width = 1.5in, height = 0.8in, keepaspectratio]{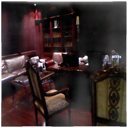} &
        \includegraphics[width = 1.5in, height = 0.8in, keepaspectratio]{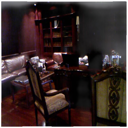} &
    	\includegraphics[width = 1.5in, height = 0.8in, keepaspectratio]{cvpr2019AuthorKit/latex/nyu_depth/gt/23.png} \\
      	Input & L1 & L1 $\xrightarrow{}$ MSE & L1 $\xrightarrow{}$ SSIM & MSE & MSE $\xrightarrow{}$ L1 & MSE $\xrightarrow{}$ SSIM & Truth
    \end{tabular}
    \caption{Examples of images produced using different loss functions and our Small FastNet architecture. Each column shows results for a different loss function. Images are from the NYU Depth dataset.}
\label{figure:lossimages}
\end{figure*}

We investigated the impact of loss function selection when optimizing our model on the NYU Depth dataset. Specifically, we fitted our models using a least absolute deviations (L1) loss and MSE loss baseline, and then further trained with a second refinement loss function. Refinement functions considered were: content loss \cite{DBLP:journals/corr/JohnsonAL16}, L1 loss, MSE loss, and SSIM loss. For the purpose of training time, we trained with our smallest model, Small FastNet. Results from this study are summarized in Table \ref{table:funcs}. Results indicate that training with L1 loss followed by MSE refinement generates images with the highest PSNR, whereas training with MSE loss followed by SSIM refinement generates images with the highest SSIM. Images generated with any of the loss functions studied are qualitatively similar, as shown in Figure \ref{figure:lossimages}.  

\begin{table}[h] 
\centering
\begin{tabular}{p{0.45\linewidth}p{0.15\linewidth}p{0.15\linewidth}}
\hline
 Loss Function & PSNR & SSIM \\
\hline
\hline
L1 & 26.34 & 0.9324 \\
L1 $\xrightarrow{}$ MSE &  \textbf{26.74} &  0.9358 \\
L1 $\xrightarrow{}$ SSIM & 25.68  & 0.9364 \\
MSE & 24.08 & 0.9034 \\
MSE $\xrightarrow{}$ L1  &  25.81 &  0.9272 \\
MSE $\xrightarrow{}$ SSIM  & 25.18 & \textbf{0.9439} \\
MSE $\xrightarrow{}$ Content Loss & 22.12 & 0.8559 \\
\hline
\end{tabular}
\caption{Comparison of loss functions used to train Small FastNet. Models were trained and evaluated on the NYU Depth dataset.}
\label{table:funcs}
\end{table}

\subsection{Timing Benchmarks}
We performed timing benchmarks to help asses the feasibility of introducing our method as a pre-processing step for computer vision algorithms in real-time systems. The average timing over 20 runs is presented on both the Titan RTX desktop GPU and Tegra Xavier edge GPU. We progressively increased input resolution until we could no longer process a given input batch size due to GPU memory limitations. Timing results are given in frames per second for both floating point 32 and floating point 16. Full timing results are presented in Table \ref{table:timing}. Unsurprisingly, the biggest timing gains come from utilizing a batch size greater than 1 and operating at floating point 16. For real-time applications, this introduces latency in exchange for throughput.  

\subsection{Comparison with State-of-the-Art Methods}
\subsubsection{Results on NYU Depth Dataset}
For the NYU Depth dataset, we show state-of-the-art performance using our Big FastNet model trained with MSE loss and SSIM loss as refinement. Additionally, the model performs efficiently relative to its parameter count. Our model width can also be scaled down in exchange for SSIM and PSNR. For instance, Small FastNet has 11 million parameters, 6x smaller than Zhang and Patel's \cite{dehaze_zhang_2018} approach, and still performs competitively. Results for our method and other approaches are summarized in Table \ref{table:NYU}.
\begin{table}[!ht]
\small
\centering
\begin{tabular}{p{0.48\linewidth}p{0.1\linewidth}p{0.1\linewidth}p{0.15\linewidth}}
\hline
Model & PSNR & SSIM & Parameters \\
\hline
\hline
Input & 13.85 & 0.70 & -\\ 
He. \etal\cite{he2011single}(CVPR'09)& - & 0.86 & - \\
Zhu. \etal\cite{zhu2015fast}(TIP'15)& - & 0.86 & - \\
Ren. \etal\cite{ren2016single}(ECCV'16) & - & 0.82 & \textbf{0.0084} \\ 
Berman. \etal\cite{NonLocalImageDehazing}(CVPR'16) & 16.92 & 0.80 & - \\
Li. \etal\cite{Li_2017_ICCV}(ICCV'17) & - & 0.88 & 0.018\\
Small FastNet (Ours) & 25.18 & 0.94 & 11.55 \\
Zhang \& Patel \cite{dehaze_zhang_2018}(CVPR'18) & 29.28 & 0.96 & 66.89\\
\textbf{Big FastNet (Ours)} & \textbf{30.37} & \textbf{0.97} & 28.78 \\
\hline
\end{tabular}
\caption{Quantitative comparison of our Big FastNet method with other state-of-the-art approaches tested on the NYU Depth test dataset. SSIM of other methods are drawn from \cite{dehaze_zhang_2018} while PSNR values and parameter counts were reproduced with the original public implementations. Parameter count is measured in millions. While smaller models exist, our method has fewer parameters and higher SSIM and PSNR than current state-of-the-art methods.}
\label{table:NYU}
\end{table}
\subsubsection{Results on High-Resolution O/I-HAZE Dataset}
We evaluated our method on the benchmark high-resolution O/I-HAZE datasets \cite{I-HAZE_2018,O-HAZE_2018}. Because of limitations in GPU memory, we used our Small FastNet model in this evaluation. Because this model has fewer parameters than other models studied, we were able to perform inference on the native full-resolution test imagery with a Titan RTX GPU. Past approaches typically use one of two methods: (1) forward pass patches of the test image and stitch the final output, or (2) operate on lower-input resolution imagery and rescale the output \cite{ancuti2018ntire}. Two models were trained separately using MSE loss, one on the O-HAZE dataset and one on the I-HAZE dataset. Each model was trained using the NYU Depth dataset pretrained model generated in earlier experiments rather than from randomly initialized weights. Training loss converged after only a few epochs, indicating that features learned from the NYU Depth dataset transfer well to other datasets, such as the O/I-HAZE datasets. To train our models, we augmented the dataset by extracting multi-scale patches reshaped to our training input size.\par
Our Small FastNet results are competitive with results from the 2018 NTIRE Image Dehazing Challenge, Indoor and Outdoor tracks \cite{ancuti2018ntire}. We achieve \textbf{SSIM of 0.8089 and PSNR of 18.56 for the I-HAZE test dataset} and \textbf{SSIM of 0.7459 and PSNR of 22.07 for the O-HAZE test dataset}. Each metric is ranked within the top 10 for its category with respect to the 2018 NTIRE Image Dehazing Challenge \cite{ancuti2018ntire}.\par
\begin{figure}[h!]
\centering
\setlength\tabcolsep{2pt}
\begin{tabular}{ccc}
\includegraphics[width = 1in]{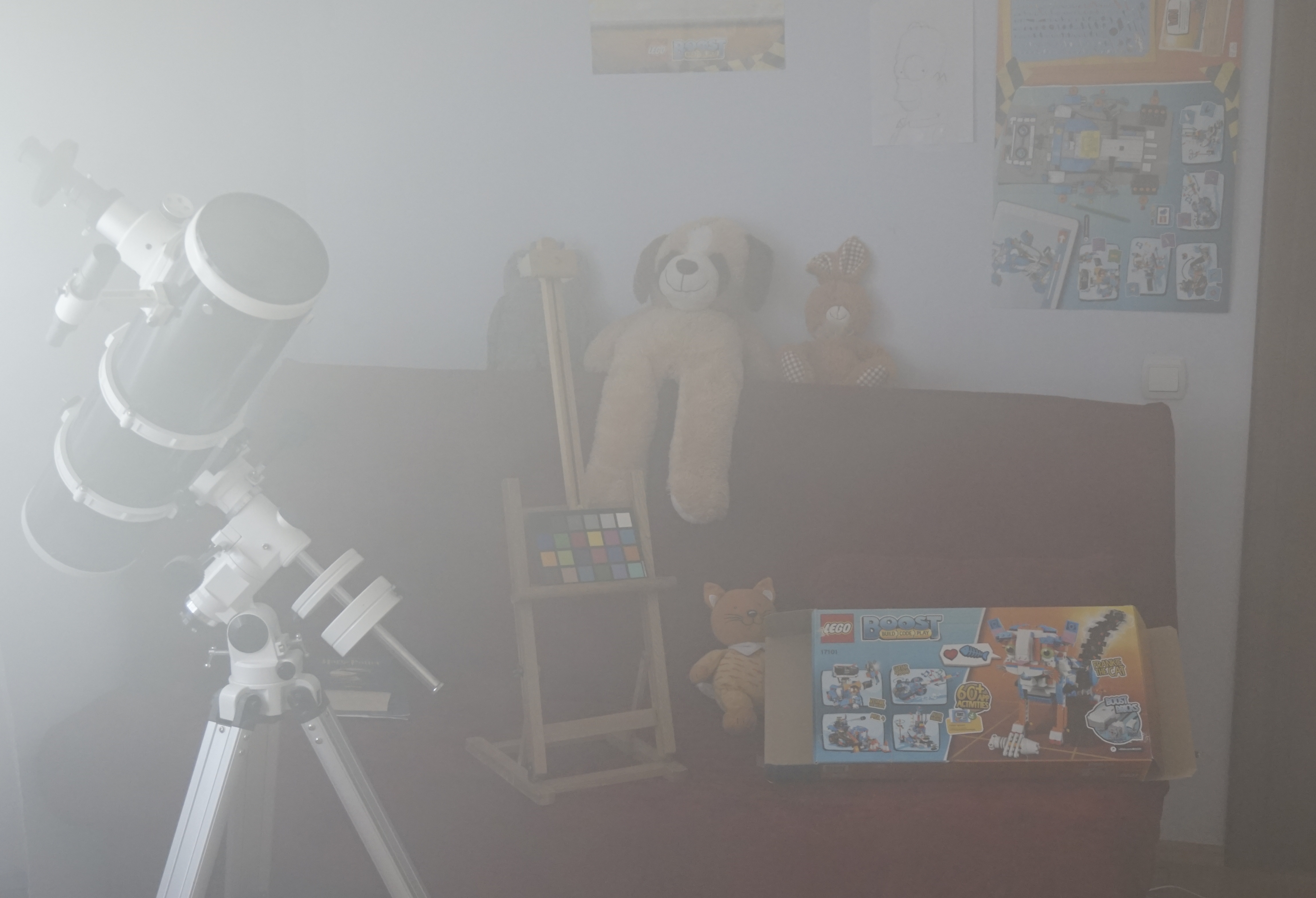} &
\includegraphics[width = 1in]{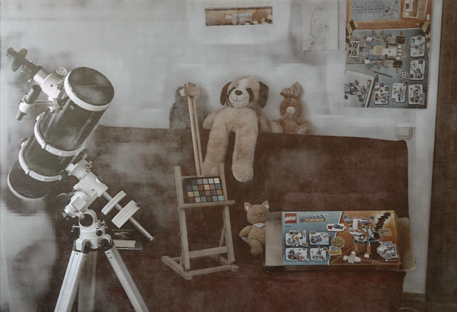} & 
\includegraphics[width = 1in]{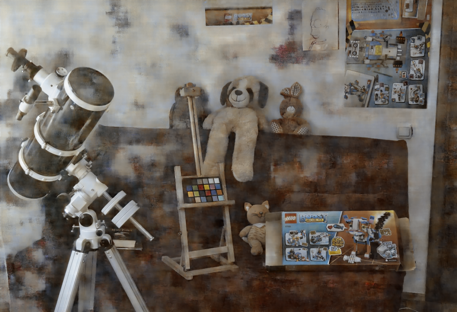} \\
\includegraphics[width = 1in]{cvpr2019AuthorKit/latex/io_haze/44_outdoor_haze.png} &
\includegraphics[width = 1in]{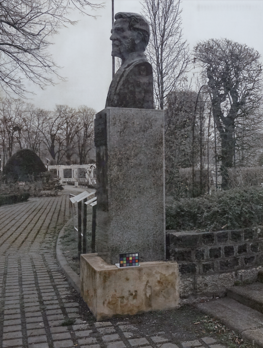} & 
\includegraphics[width = 1in]{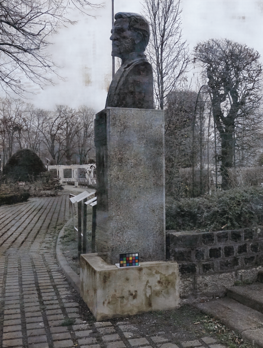} \\
\footnotesize{Input} & \footnotesize{Early Training Output} & \footnotesize{Top SSIM Output}
\end{tabular}
\caption{Example of images produced by our Small FastNet architecture for the I-HAZE dataset (top) and O-HAZE dataset (bottom). Overfitting leads to improved SSIM and PSNR, but causes qualitative defects.}
\label{3}
\end{figure}

Figure \ref{3} shows several images generated with our approach, including results from early training and results from the end of training when top SSIM has been reached. Although SSIM and PSNR continue to improve in later epochs of training, artifacts in imagery commonly seen in neural network approaches for image generation become noticeable. This indicates that it is important to not only maximize SSIM and PSNR, but also to conduct thorough qualitative analysis when evaluating top models for image dehazing. Because pixel values within areas of continuous dense haze are likely unrecoverable, the neural network learns to minimize its loss by using the average pixel value learned from similar areas in training data when it encounters dense haze. This causes the artifacts observed in Figure \ref{3}. In short, areas with unrecoverable pixel values are substituted with random training artifacts, which are likely to be strong indicators of overfitting.

\begin{table*}[t]
\centering
\begin{tabular}{p{0.08\linewidth}p{0.05\linewidth}p{0.08\linewidth}p{0.08\linewidth}p{0.07\linewidth}p{0.07\linewidth}p{0.08\linewidth}p{0.08\linewidth}p{0.07\linewidth}p{0.07\linewidth}}
\hline
Image Size & Batch Size & Small-32 RTX & Small-16 RTX & Big-32 RTX & Big-16 RTX & Small-32 Xavier & Small-16 Xavier & Big-32 Xavier & Big-16 Xavier  \\
\hline
\hline
512x512 & 1 & 190.35 & 164.17 & 97.44 & 131.59 & 16.31 & 23.07 & 8.41 & 17.46 \\
512x512 & 4 & 225.98 & 313.38 & 132.91 & 231.41 & 16.21 & 28.64 & 8.46 & 20.80 \\
512x512 & 8 & 231.35 & 367.20 & 138.07 & 263.53 & 16.08 & 29.09 & 8.42 & 21.12 \\[.07cm]
1024x1024 & 1 & 58.03 & 79.21 & 33.15 & 57.39 & 3.92 & 7.54 & 2.02 & 5.18 \\
1024x1024 & 4 & 60.11 & 100.43 & 34.66 & 70.34 & 3.77 & 7.83 & 1.98 & 5.29 \\[.07cm]
2048x2048 & 1 & 14.65 & 25.31 & 8.45 & 17.53 & 0.87 & 1.95 & 0.48 & 1.28 \\
\hline
\end{tabular}
\caption{Timing benchmarks generated with TensorRT to assess the feasibility of introducing our model to a real-time system. In general, increasing batch size allows for higher frames per second (FPS) processing in exchange for latency. Small or Big indicates the FastNet model used. The floating point precision is indicated by 16 or 32. Results are described in average FPS over 20 iterations. }
\label{table:timing}
\end{table*}
\subsubsection{Results on Dense-Haze Dataset}
The 2019 NTIRE Image Dehazing Challenge \cite{NTIRE_Dehazing_2019} introduces a novel dataset, called Dense-Haze, containing challenging, dense haze imagery. The dataset includes 45 training images, 5 validation images, and 5 test images, each with a resolution of 1600 x 1200 pixels. We trained our DualFastNet model on all 45 training images, as well as 2 validation images, leaving 3 images for validation and 5 for testing. Training started with randomly initialized weights and data were randomly cropped and rotated throughout. Our model produced results that were competitive with other models in the challenge, achieving a PSNR score of \textbf{16.37} and an SSIM score of \textbf{0.569} on test images. Examples of the images generated are shown in Figure \ref{figure:preview}.\par
Although Big FastNet outperforms our other models in earlier experiments, these experiments did not use models trained on sparse datasets with randomly initialized weights as was done in this challenge. We have observed that when limited to fewer training samples, DualFastNet can generate superior results, indicating that the atmospheric scattering model is a helpful prior in certain conditions. Specifically, on the Dense-Haze dataset, Big FastNet achieved the highest SSIM and DualFastNet achieved the highest PSNR. A qualitative study of output images was done that informed the decision to use DualFastNet in our challenge submission. 

\section{Conclusion}
In this paper, we propose a family of novel neural network architectures for single image dehazing, as well as present both quantitative and a qualitative evaluation of these architectures and their loss functions. On the NYU Depth dataset, Big FastNet, our largest model, outperforms its smaller variant and our architecture based on the atmospheric scattering model. Additionally, this approach outperforms other state-of-the-art neural networks on the NYU Depth dataset in both performance and efficiency. However, our experimental results indicate that the atmospheric scattering model is a useful prior for a neural network architecture when training data is limited. Our architectures can be run as part of real-time systems on edge GPUs and have been benchmarked on multiple input imagery sizes. Finally, we discuss our results and challenges working with the O-HAZE, I-HAZE, and Dense-Haze datasets. In the 2019 NTIRE Image Dehazing Challenge, our efficient, atmospheric scattering model-based neural network architecture, DualFastNet, achieved competitive results, obtaining a PSNR score of \textbf{16.37} and an SSIM score of \textbf{0.569} on test images.
\section*{Acknowledgements}
DISTRIBUTION STATEMENT A. Approved for public release. Distribution is unlimited.
This material is based upon work supported under Air Force Contract No. FA8702-15-D-0001. Any opinions, findings, conclusions or recommendations expressed in this
material are those of the author(s) and do not necessarily reflect the views of the U.S. Air Force.
{\small
\bibliographystyle{ieee_fullname}
\bibliography{NTIRE2019_CameraReady}
}

\end{document}